# Deep Learning vs. Human Graders for Classifying Severity Levels of Diabetic Retinopathy in a Real-World Nationwide Screening Program

Paisan Raumviboonsuk, MD[1*]
Jonathan Krause, PhD[2*]
Peranut Chotcomwongse, MD[1†]
Rory Sayres, PhD[2†]
Rajiv Raman, MD[3]
Kasumi Widner[2]
Bilson J L Campana, PhD[2]
Sonia Phene[2]
Kornwipa Hemarat, MD[4]
Mongkol Tadarati, MD[1]
Sukhum Silpa-Acha, MD[1]
Jirawut Limwattanayingyong, MD[1]
Chetan Rao, MD[3]
Oscar Kuruvilla, MD[5]
Jesse Jung, MD[6]
Jeffrey Tan, MD[7]
Surapong Orprayoon, MD[8]
Chawawat Kangwanwongpaisan, MD[9]
Ramase Sukulmalpaiboon, MD[10]
Chainarong Luengchaichawang, MD[11]
Jitumporn Fuangkaew, MD[12]
Pipat Kongsap, MD[13]
Lamyong Chualinpha, RN[14]
Sarawuth Saree, MD[15]
Srirat Kawinpanitan, RN[16]
Korntip Mitvongsa, RN[17]
Siriporn Lawanasakol, RN[18]
Chaiyasit Thepchatri, MD[19]
Lalita Wongpichedchai, MBBS, MCTM, FM, Oph[20]
Greg S Corrado, PhD[2]
Lily Peng, MD, PhD[2‡**]
Dale R Webster, PhD[2‡]

[1]Department of Ophthalmology, Rajavithi Hospital, Bangkok, Thailand
[2]Google AI, Google, Mountain View, CA, USA
[3]Shri Bhagwan Mahavir Vitreoretinal Services, Sankara Nethralaya, Chennai, Tamil Nadu, India
[4]Department of Ophthalmology, Vajira Hospital, Bangkok, Thailand
[5]Eye and Laser Center, Charlotte, NC, USA
[6]East Bay Retina Consultants, Oakland, CA; Department of Ophthalmology, University of California San Francisco, San Francisco, CA, USA
[7]Retina-Vitreous Associates Medical Group; Department of Ophthalmology, University of Southern California, Los Angeles, CA, USA
[8]Department of Ophthalmology, Lamphun Hospital, Lamphun, Thailand
[9]Department of Ophthalmology, Somdejphrajaotaksin Maharaj Hospital, Tak, Thailand
[10]Department of Ophthalmology, Sawanpracharak Hospital, Nakhon Sawan, Thailand
[11]Department of Ophthalmology, Nakhon Nayok Hospital, Nakhon Nayok, Thailand
[12]Department of Ophthalmology, Photharam Hospital, Ratchaburi, Thailand
[13]Department of Ophthalmology, Prapokklao Hospital, Chanthaburi, Thailand
[14]Department of Ophthalmology, Mahasarakham Hospital, Mahasarakham, Thailand
[15]Department of Ophthalmology, Nongbualamphu Hospital, Nongbualamphu, Thailand
[16]Department of Ophthalmology, Pakchongnana Hospital, Nakhon Ratchasima, Thailand
[17]Department of Ophthalmology, Mukdahan Hospital, Mukdahan, Thailand
[18]Department of Ophthalmology, Suratthani Hospital, Suratthani, Thailand
[19]Department of Ophthalmology, Sungaikolok Hospital, Narathiwat, Thailand
[20]Bangkok Metropolitan Administration Public Health Center 7, Bangkok, Thailand
[*]Equal Contribution; [†]Equal Contribution; [‡]Equal Contribution

[**]Corresponding Author
Lily Peng, MD, PhD
Google AI
1600 Amphitheatre Way
Mountain View, CA 94043
lhpeng@google.com

**Abstract**

**Objective**: Deep learning algorithms have been used to detect diabetic retinopathy (DR) with specialist-level accuracy. This study aims to validate one such algorithm on a large-scale clinical population, and compare the algorithm performance with that of human graders.

**Research Design and Methods:** 25,326 gradable retinal images of patients with diabetes from the community-based, nation-wide screening program of DR in Thailand were analyzed for DR severity and referable diabetic macular edema (DME). Grades adjudicated by a panel of international retinal specialists served as the reference standard.

**Results**: Relative to human graders, for detecting referable DR (moderate NPDR or worse), the deep learning algorithm had significantly higher sensitivity (0.97 vs. 0.74, $p<0.001$), and a slightly lower specificity (0.96 vs. 0.98, $p<0.001$). Higher sensitivity of the algorithm was also observed for each of the categories of severe or worse DR, PDR, and DME ($p<0.001$ for all comparisons). The quadratic-weighted kappa for determination of DR severity levels by the algorithm and human graders was 0.85 and 0.78 respectively ($p<0.001$ for the difference).

**Conclusions**: Across different severity levels of DR for determining referable disease, deep learning significantly reduced the false negative rate (by 23%) at the cost of slightly higher false positive rates (2%). Deep learning algorithms may serve as a valuable tool for DR screening.

**Introduction**

Deep learning (DL) is a field of artificial intelligence which has been applied to develop algorithms for the detection of diabetic retinopathy (DR) with high (>90%) sensitivity and specificity for referable disease (moderate non-proliferative diabetic retinopathy (NPDR) or worse) (1–3). In addition to high screening accuracy, DL also has advantages in terms of resource consumption, consistency, and scalability, and has the potential to be deployed as an alternative to human graders for classifying or triaging retinal photographs in DR screening programs.

In Thailand, there are 1,500 ophthalmologists, including 200 retinal specialists, who provide ophthalmic care to approximately 4.5 million patients with diabetes. Half of the ophthalmologists and retinal specialists practice in Bangkok, the capital of the country, while a majority of patients with diabetes live in areas 100 kilometers or more from provincial hospitals, where ophthalmologists typically practice. The latest Thailand National Survey of Blindness conducted in 2006-2007 (4) showed that 34% of patients with diabetes had low vision or blindness in either eye. DR was and continues to be the most common retinal disease that causes bilateral low vision (4,5).

A national screening program for DR was set up by the Ministry of Public Health of Thailand in 2013. The screening was conducted in each of the 13 health regions with an initial target of screening at least 60% of diabetic patients in each region. Unfortunately, Ministry data indicates that less than 50% of the diabetic patients were screened every year since the inception of the program. Because this was in part due to the lack of trained graders, deploying DL in the screening program for DR in Thailand has the potential to solve some of these problems (6). Similar issues have been observed in the United Kingdom (7).

Several DL algorithms for DR have shown promise in populations with multiethnic diabetic patients (1–3). However, before the deployment of DL for screening DR, additional large-scale validation on screening populations that are distinct from the original developmental datasets will be critical. In addition, the use of rigorous reference standards that are adjudicated by retinal specialists is important for robust evaluation of the algorithm and human graders (2). Lastly, the diagnostic accuracy of DL algorithms should be compared with human graders who routinely grade retinal images in a screening population.

This study was conducted to assess the screening performance of the DL algorithm compared to real-world graders for classifying multiple clinically relevant severity levels of DR in the national screening program for DR in Thailand.

**Research Design & Methods**

This study was approved by the Ethical Review Committee for Research in Human Subjects of the Ministry of Public Health of Thailand and the Ethical Committees of hospitals or health centers from which retinal images of patients with diabetes were used. Patients gave informed consents allowing their retinal images to be used for research. This study was registered in the Thai Clinical Trials Registry, Registration Number TCTR20180716003.

<u>Data acquisition</u>

Diabetic patients were randomly identified from a national registry of diabetic patients, representing hospitals or health centers in each of the 13 health regions in Thailand. Patients were included if they had fundus images of either eye captured using retinal cameras in both the years 2015 and 2017, as part of a 2-year longitudinal study on DR. Retinal images of the patients were single-field, 45-degree field of view, and contained the optic disc and macula, centered on the macula. A variety of cameras were used for image acquisition including ones manufactured by 3nethra, Canon, Kowa, Nidek, Topcon, and Zeiss (Supplemental Table S6). Images were retrieved from the digital archives from retinal cameras utilized in the Thailand DR national screening program. Images were excluded from analysis if they were from patients who had other retinal diseases that precluded classification of severity of DR or DME, such as age-related macular degeneration (AMD) and other retinal vascular diseases.

Definition of DR severity levels and DME

Severity levels of DR and DME were defined according to the International Clinical Classification of DR (ICDR) disease severity scale (8). In short, DR was classified into no DR, mild non-proliferative DR (NPDR), moderate NPDR, severe NPDR, and proliferative DR (PDR). DME was identified as referable DME when hard exudates were found within the distance of 1 disc-diameter from the center of the fovea (9,10).

Sample size estimation

According to previous community-based studies of DR in Thailand (11), the prevalence of sight-threatening DR (PDR, severe NPDR, or DME) was approximately 6% of patients with diabetes. With a margin of error of 10%, type 1 error at 0.5 and type 2 error at 0.2, the sample size was estimated at no less than 6,112 patients with diabetes. A rate of ungradable images at 20% was estimated, therefore, at least 7,450 patients with diabetes were required. The distribution of diabetic patients included from each region was in proportion with the distribution of diabetic patients form each region. The numbers of patients from each of the 13 regions are listed in Table 1.

Deep learning algorithm

The development of the deep learning algorithm for predicting DR and DME is described in detail in Krause *et al*. (2) Briefly, a convolutional neural network was trained with an "Inception-v4" (12) architecture that predicted a 5-point DR grade, referable DME, gradability of both DR and DME, and an overall image quality score. The input to the neural network was a fundus image with a resolution of 779 x 779 pixels. Through the use of many stages of computation, parameterized by millions of numbers, the network outputs a real-valued number between 0.0 and 1.0 for each prediction, indicating its confidence. During training, the model was given different images from the training set with a known severity rating for DR, and the model predicted its confidence in each severity level of DR, slowly adjusting its parameters over the course of the training process to increase its accuracy. The model was evaluated on a tuning dataset throughout the training process, which was used to determine model hyperparameters. An "ensemble" of ten individual models was then created to combine their predictions for the final output. To turn the model's confidence-based outputs into discrete predictions, a threshold on the confidence was used for each binary output (DME, DR Gradability, and DME Gradability), and a cascade of thresholds (2) was used to output a single DR Severity level.

Grading by regional graders

The DL algorithm was compared to 13 human regional graders who actually grade retinal images for the screening program in each of the 13 health regions. In this study, each grader only graded images that have been screened in his or her own region. Some of the graders were general ophthalmologists and others were trained ophthalmic nurses or technicians. Each grader received standard grading protocols for DR and DME, including instructions for the web-based grading tool before the commencement of the study, and each was required to use the same web-based tool for online grading of the retinal images. A tutorial session was conducted for all the graders before the commencement of grading.

Reference standard

There were 2 groups of retinal specialists who graded retinal images for the reference standard. One group was assigned to grade for DR severity level and another for referable DME.

For gradability, a subset of ~1000 images each for DR and DME where the regional grader disagreed with the DL algorithm on image gradability underwent adjudication. For the remainder of the analysis for both DR and DME, images graded as ungradable by either the algorithm or regional grader were excluded. Additionally, for images which were not adjudicated but the DL algorithm and regional graders were in agreement about the severity of DR, the agreed-upon grade was used as the reference standard. A similar rule was applied for analysis of DME.

      For grading DR severity levels the ICDR scale was used. In order to reduce adjudication time for quality of grading, adjudicators were assigned to grade a subset of the images. This subset included all images for which the regional grader and the DL algorithm were in disagreement and at least one graded as moderate NPDR or worse; a random sample of 75 images for which the algorithm and regional grader were in agreement and graded as moderate NDPR or worse; and a random sample of 1,175 images for which the algorithm and regional grader both graded as less than moderate NPDR. This random sample represented 5% of all images with agreement between the two modalities. The ratio of images with moderate NPDR or worse to those with less than moderate NPDR in this random sample was proportional to the entire population. For the purposes of this study, no DR and mild DR were considered a single category, and only images that both the algorithm and regional grader deemed gradable for DR were adjudicated. For grading referable DME, retinal specialists were assigned to grade all images for which the regional graders and DL were in disagreement about the binary presence or absence of DME, and 5% of the rest of the images were randomly assigned from the subset of those images also determined to be gradable for DME by both the regional grader and the algorithm. Most of the disagreement occurred in cases that were graded as moderate or worse DR or referable DME. Further review of the discrepancies revealed that most of the cases were fairly ambiguous and adjudicators tended to err on the side of increased sensitivity for a screening setting. Overall, for moderate or worse DR, the regional grader and algorithm grade agreed with adjudication 96.3% of the time. For DME, the agreement rate was 97.1%.

The retinal specialists that served as the reference standard in this study were from Thailand, India, and the United States. There were two retinal specialists per group. Each group graded the images independently, and the same instructions for web-based grading that the regional graders used were issued to them before grading. They were also required to use the same web-based grading tool as the regional graders. In addition, the group that graded DR severity level had a teleconference and graded a small set of images together to ensure congruence.

The adjudication process was as follows: First, both retinal specialists in a group independently graded each image. Then, until consensus was reached, the retinal specialists took turns revising their grades, each time with access to their previous grade and the other retinal specialist's grade, as well as any additional comments about the case either retinal specialist left. If there was still disagreement after each grader had graded the image three times in this way, then the image's reference standard was determined independently by a separate, senior retinal specialist. For grading DR severity levels, differences between no DR and mild NPDR were not adjudicated in order to focus adjudication time on referable disease.

Statistical Analysis

Primary metrics assessed were sensitivity, specificity, and area under the receiver operating characteristic curve (AUC). Confidence intervals for sensitivities and specificities were calculated using a Clopper-Pearson interval, and confidence intervals for the quadratic-weighted Cohen's kappa was calculated using a bootstrap, both calculated at the 95% level. All P-values were calculated using two-sided permutation tests.

**Results**

The characteristics of the images and patients included in this study are described in Table 1. This cohort consisted from 7,517 patients, of which 67.5% were women. The average age was 61.13 (SD = 10.96).

Out of 29,943 images, 4,595 were deemed not gradable for DR by either the regional grader, the DL algorithm, or both (Supplemental Table S1). A sample of images where the regional grader disagreed with the DL algorithm on image gradability underwent adjudication and the results of the adjudication are presented in the Supplemental Tables S2 and S3. Adjudicators were approximately 2.5 times more likely to agree with the algorithm than regional graders about the gradability of images for DR. For this difficult image subset, adjudicators agreed with the regional grader 28.5% of the time vs 71.5% of the time with the algorithm. For DME gradability, they were just as likely to agree with the algorithm as the regional graders.

A comparison of the performance of regional graders in Thailand and the algorithm compared to the reference standard for all gradable images is summarized in Supplemental Table S4. Out of all of the gradable images, the composite sensitivity (i.e. generated by pooling all of the patients and then computing the metric) of the graders for detecting moderate or worse DR was 0.740 (ranging from 0.413-0.954 across regional graders) and the specificity was 0.982 (range: 0.943-1.000). The DL algorithm had a sensitivity of 0.970 (range: 0.887-0.993), specificity of 0.957 (range: 0.903-0.987) and AUC of 0.988 (range: 0.978-0.996) (Figure 1). These differences in sensitivity (23%) and specificity (-2.5%) between the algorithm and regional graders were statistically significant ($p<0.001$). The algorithm also showed performance better than or equal to that of composite grading of regional graders for severe or worse DR and PDR, with AUC values of 0.991 (range: 0.978-0.997) and 0.994 (range: 0.974-1.00), respectively (Figure 1). Using moderate or worse DR as the threshold for referral for regional graders would result in a sensitivity of 0.917 (95% CI 0.870-0.951) for severe or worse DR. For the algorithm, this would correspond to a sensitivity of 0.990 (95% CI: 0.965-0.999). For PDR, the graders would have a sensitivity of 0.859 (95% CI: 0.821-0.892) and the algorithm would have a sensitivity of 0.955 (95% CI: 0.929-0.973).

Results for DME were similar. The sensitivity for regional graders for detecting referable DME was 0.613 (range: 0.421-0.805 across regions), and the specificity was 0.992 (range: 0.973-0.999). For the DL algorithm, sensitivity was measured at 0.940 (range: 0.847-1.000), specificity was measured at 0.982 (range: 0.947-0.992), and AUC of 0.993 (range: 0.981-0.998; 95% CI 0.991-0.994).

Because more rapid referral is warranted for cases with severe or worse DR and/or DME, the performance of the algorithm for these cases was also examined. At this threshold, regional graders had a sensitivity of 0.603 (range: 0.308-0.782), specificity of 0.997 (range: 0.993-0.999), while the algorithm had a sensitivity of 0.927 (range: 0.810-1.000) and specificity of 0.976 (range: 0.915-0.993). Using PDR and/or DME as the threshold yields similar performance metrics as severe or worse DR and/or DME because the number of DME cases outnumbers that of PDR cases (Figure S1). Additional results for using individual DR severity levels as the threshold are summarized in Supplemental Figure S1.

The performance of the graders cannot be directly compared to each other because they each graded a different set of images that correspond to their region. However, the algorithm's performance for images from each region could be compared directly to the regional grader from that region (Figure 2). In nearly all regions, the algorithm's sensitivity was significantly higher than that of the respective regional grader for moderate or worse DR and for DME. In region 5, the algorithm's sensitivity was lower than that of the regional grader for moderate or worse DR, but the difference was not statistically significant ($p=0.83$).

The overall agreement between the regional graders and algorithm in comparison to the reference standard for DR and DME is shown in a confusion matrix presented in Table 2. Furthermore, to compare the agreement for the entire range of DR severities (No/Mild, Moderate, Severe, and Proliferative) and for each region, quadratic-weighted Cohen's kappa was used (Supplemental Table S5). Regional graders were measured at 0.728 (range: 0.504-0.853 across regions) and 0.776 (range: 0.663-0.878), $p<0.001$ for the difference.

While the output of the algorithm is ultimately distilled in a categorical call (e.g. severe NPDR vs PDR), the algorithm originally returns a value between 0 and 1 for each level of DR and DME, indicating a confidence for each severity level. Analysis of the model's performance based on the maximum score of both the DR and DME predictions showed that the algorithm was more sensitive than regional graders at all ranges of confidence score. However, when the algorithm was uncertain (maximum score < 0.7), the specificity of the algorithm was much lower than that of the regional grader at this particular operating point (Figure S2).

**Conclusions**

This study represents one of the largest clinical validations of a deep learning algorithm in a population that is distinct from which the algorithm was trained. In addition, this external validation was conducted in direct comparison with the actual graders in the screening program of the same population. This is pivotal since many countries in the world adopted trained graders for their screening programs for DR. These include the UK (13), Malaysia (14), South Africa (15), among others. Furthermore, according to the statement by American Academy of Ophthalmology (AAO) (15,16), there is a strong (level 1) evidence that single-field retinal photography with interpretation by trained graders can serve as a screening tool to identify patients with DR for referral for ophthalmic evaluation and management.

Most algorithms in previous studies of DL for DR screening have simplified the various levels of DR into binary predictions, either referable and non-referable or with and without sight-threatening DR. However, in real-world situations patient management can be different at each level of DR. For example, patients with PDR may require higher urgency for referral for panretinal photocoagulation or intravitreal injections compared to another severity level, such as severe NPDR without diabetic macular edema (DME). On the other hand, patients with moderate NPDR without DME, although labelled as referable, may not require treatment but still require periodic close monitoring by ophthalmologists or retinal specialists. Identifying the referable group of patients requiring treatment accurately may save community resources. Validation of the performance of DL for classifying severity levels of DR would therefore be essential for real-world screening of DR.

Prior to this study, one of the largest head-to-head comparisons between deep learning and human graders was performed by Ting *et al.* The study included 10 secondary validation sets (e.g. validation sets that were drawn from a population that is distinct from the one in which the model was trained), the largest consisting of over 15,000 images. The algorithm validated in this study build upon this body of work by not only showed excellent generalization on a national level across different cameras (Supplemental Table S6) and graders, but also high accuracy when measured on both binary and multi-class tasks. Grading on a more granular 5-point grade is advantageous, especially on a global scale, where follow up and management guidelines amongst the many different guidelining bodies may vary at each of the 5 levels and in the presence of possible macular edema (17). For example, while the follow up recommended by the AAO (15,16,18) can be up to 12 months for moderate NPDR with no macular edema, that recommendation changes to 3-6 months if there is macular edema and 1 month when there is clinically significant macular edema (16).

The threshold level for referral in a screening program for DR may be dependent on the resources of the program. In a lower resource setting where ophthalmologists only see severe cases, the referral threshold may be higher. In a higher resource setting where ophthalmologists prefer to see mild cases, the referral threshold may be lower. There has been some work in the literature to address the importance of this 5-severity levels grading of DR and an adjudicated reference standard (8). However, this work was not validated on a dataset from a different population until this present study. External validation of a deep learning system for accuracy of a 5-point grade should not only give an advantage for selecting an appropriate threshold with acceptable accuracy for a screening system but also act as a feedback loop to tune the accuracy of the deep learning system itself.

This study could be further improved upon. First, adjudication was performed only for moderate or worse cases. For patients with no retinopathy, screening every 2 years may be appropriate while those with mild NPDR should be screened once a year (19). Therefore, future studies should include adjudication for cases where there is disagreement between no and mild DR. Furthermore, the algorithm's performance could be improved upon for difficult cases. For example, the sensitivity for correctly detecting PDR of our deep learning algorithm was only 0.719 in this study. Even though the algorithm did not miss cases with active neovascularization but rather primarily missed cases with inactive fibrous tissue without neovascularization at the optic disc, additional training could help improve the algorithm for these cases in the future. In addition, in a real-life screening setting, grading may be performed with a combination of automated and manual grading. A preliminary analysis was performed to look at the relationship between algorithm confidence and performance of both algorithm and manual grading. Future studies could further explore how to combine the algorithm and manual grading to achieve better performance than either alone, while minimizing manual grading workload. In addition, the reference standard for DME in this study was based on monoscopic fundus photos. For DME, OCT is now considered the clinical standard, and incorporating this in the reference standard would be ideal (20). Lastly, DR screening programs generally also refer patients at high suspicion for other non-DR eye diseases such as AMD or glaucoma. The ability to detect other eye diseases would further increase the utility of these algorithms.

    This study represents an early milestone in the implementation of a DL algorithm in a large scale DR screening program. The demonstration of the algorithm's performance and generalizability compared to actual graders in a screening program lays the groundwork for other prospective studies -- to further validate the algorithm's performance in real screening workflows and to study its impact on DR screening as a whole. It is possible that, due to human nature, graders are generally more reluctant to make a commitment for diagnosis, compared with the

algorithm; this may have made their gradings more specific but less sensitive. While it is critical that a DL algorithm is accurate, it is equally important to study how the algorithm may affect clinical workflow and outcomes of patients, such as clinician and patient satisfaction, patient adherence to follow-up recommendations, and ultimately impact on disease prevention, progression and outcomes.

**Acknowledgements**

The authors would like to acknowledge the following people for their advice and assistance with this paper:

Piyada Pholnonluang, MD, Rada Kanjanaprachot, Yothin Jindaluang, MD, Nitaya Suwannaporn, Niteeya Narapisut, Naraporn Sumalgun, Tanawan Sinprasert, PhD, Tienchai Methanoppakhun, MD, Ratana Boonyong, Preeyawan Tanomsumpan, Rojsak Phuphiphat, Porntip Nitikarun, MD, Phirun Wongwaithayakonkun, MD, Achareeya Saenmee, MD, Cheunnapa Komkam, Supaporn Numsui, Praween Tantiprapha, Sopon Nilkumhang, MD, Roongruedee Thangwongchai, MD, Supaporn Petcharat, Jansiri Laiprasert, MD, Premrudee Maneechaweng, Sareerah Waesamah, Poowadon Pimpakun, Prapaphan Charncheaw, Ramida Panyarattannakul, Suwannee Saelee, Nutchana Onarun, Yun Liu, PhD, Peter Wubbels, Florence Thng, Sunny Virmani, Varun Gulshan, PhD, Philip Nelson, David Coz, Derek Wu, Elin Pedersen, PhD, William Chen, Jessica Yoshimi, Xiang Ji, Quang Duong, Brian Basham

# Tables & Figures

| Grader Type | All regions | 1 | 2 | 3 | 4 | 5 | 6 | 7 | 8 | 9 | 10 | 11 | 12 | 13 |
|---|---|---|---|---|---|---|---|---|---|---|---|---|---|---|
| Grader Type | | MD | MD | MD | MD | MD | MD | Nurse | MD | Nurse | Nurse | Nurse | Nurse | Tech |
| Total Patients | 7,517 | 100 | 620 | 569 | 440 | 513 | 620 | 680 | 1,005 | 750 | 370 | 500 | 250 | 1,000 |
| Total Images | 29,985 | 764 | 2,467 | 2,256 | 1,760 | 2,051 | 2,424 | 2,720 | 4,020 | 2,989 | 1,582 | 1,986 | 968 | 3,998 |
| % No/Mild DR | 87.83 | 68.30 | 92.32 | 94.02 | 92.95 | 87.85 | 88.81 | 82.85 | 82.10 | 89.62 | 75.24 | 92.45 | 86.67 | 93.81 |
| % Moderate NPDR | 9.80 | 23.84 | 5.65 | 5.35 | 5.17 | 7.40 | 8.77 | 12.53 | 16.21 | 8.14 | 20.06 | 6.15 | 10.75 | 5.39 |
| % Severe NPDR | 0.81 | 3.23 | 0.60 | 0.19 | 0.77 | 0.56 | 0.75 | 1.38 | 0.57 | 1.30 | 1.16 | 0.27 | 1.34 | 0.40 |
| % PDR | 1.57 | 4.63 | 1.43 | 0.43 | 1.12 | 4.18 | 1.67 | 3.24 | 1.11 | 0.94 | 3.55 | 1.14 | 1.23 | 0.40 |
| % Referable DME | 6.23 | 17.41 | 3.08 | 3.50 | 3.30 | 4.00 | 7.47 | 8.68 | 8.81 | 6.62 | 15.42 | 3.83 | 6.20 | 2.30 |
| % Female | 69 | 66 | 61 | 67 | 68 | 65 | 69 | 69 | 77 | 71 | 64 | 75 | 63 | 68 |
| % Male | 31 | 34 | 39 | 33 | 32 | 35 | 31 | 31 | 23 | 29 | 36 | 25 | 37 | 32 |
| Age | 59 (52, 66) | 58 (53, 64) | 57 (50, 63) | 59 (52, 66) | 63 (57, 70) | 62 (54, 70) | 58 (51, 65) | 62 (54, 68) | 59 (53, 66) | 56 (49, 64) | 56 (49, 64) | 63 (56, 71) | 56 (50, 61) | 59 (51, 67) |
| HbA1c (%) | 7.3 (6.5, 8.6) | 7.6 (6.9, 8.5) | 7.0 (6.4, 8.2) | 7.3 (6.4, 8.5) | 7.2 (6.5, 8.4) | 7.2 (6.3, 8.5) | 7.2 (6.5, 8.1) | 7.7 (6.8, 9.3) | 7.6 (6.5, 9.1) | 7.2 (6.3, 8.5) | 8.4 (7.3, 9.8) | 7.2 (6.5, 8.2) | 8.1 (7.1, 9.6) | 7.0 (6.3, 8.0) |
| FBS (mg/dL) | 139 (118, 169) | 130 (110, 172) | 136 (118, 168) | 138 (115, 166) | 133 (114, 156) | 150 (126, 181) | 140 (122, 175) | 140 (118, 199) | 144 (121, 172) | 133 (107, 170) | 149 (122, 188) | 131 (115, 154) | 149 (130, 186) | 136 (118, 163) |
| LDL (mg/dL) | 105 (83, 130) | 117 (107, 147) | 113 (90, 135) | 102 (79, 124) | 101 (81, 128) | 94 (75, 120) | 103 (80, 129) | 102 (93, 122) | 109 (86, 132) | 107 (86, 131) | 104 (85, 124) | 96 (73, 119) | 118 (96, 142) | 108 (88, 132) |

**Table 1.** Summary of patient characteristics, including breakdowns by region. For blood sample measures and visual acuity, values reflect the distribution across patients at first visit. MD = Ophthalmologist; n/a: Data not collected in these regions. Numeric values indicate the median across a distribution; values in parentheses indicate the 25th and 75th percentiles.

|  |  | Regional Grader Label | | | |
|---|---|---|---|---|---|
|  |  | No/Mild NPDR | Moderate NPDR | Severe NPDR | Proliferative DR |
| Reference Standard | No/Mild | 21,843 | 355 | 12 | 33 |
|  | Moderate NPDR | 729 | 1,724 | 13 | 15 |
|  | Severe NPDR | 17 | 79 | 100 | 8 |
|  | Proliferative DR | 56 | 87 | 14 | 241 |

Quadratic-Weighted Kappa: 0.776 [0.757-0.792]

|  |  | Algorithm Label | | | |
|---|---|---|---|---|---|
|  |  | No/Mild NPDR | Moderate NPDR | Severe NPDR | Proliferative DR |
| Reference Standard | No/Mild NPDR | 21,288 | 902 | 38 | 15 |
|  | Moderate NPDR | 72 | 1,866 | 525 | 18 |
|  | Severe NPDR | 2 | 6 | 192 | 4 |
|  | Proliferative DR | 18 | 18 | 76 | 286 |

Quadratic-Weighted Kappa: 0.846 [0.835-0.856]

|  |  | Regional Grader Label | |
|---|---|---|---|
|  |  | No DME | DME |
| Reference Standard | No DME | 22,182 | 168 |
|  | DME | 723 | 1,146 |

|  | | Algorithm Label | |
|---|---|---|---|
|  | | No DME | DME |
| Reference Standard | No DME | 21,951 | 399 |
|  | DME | 112 | 1757 |

**Table 2.** Agreement on the image level between the reference standard and regional graders or the algorithm for DR severity grading and DME. Adjudication was performed only for images where either the regional grader or the algorithm identified as moderate and above. Thus, for DR, non-referable cases (No/Mild) are combined into a non-referable bucket.

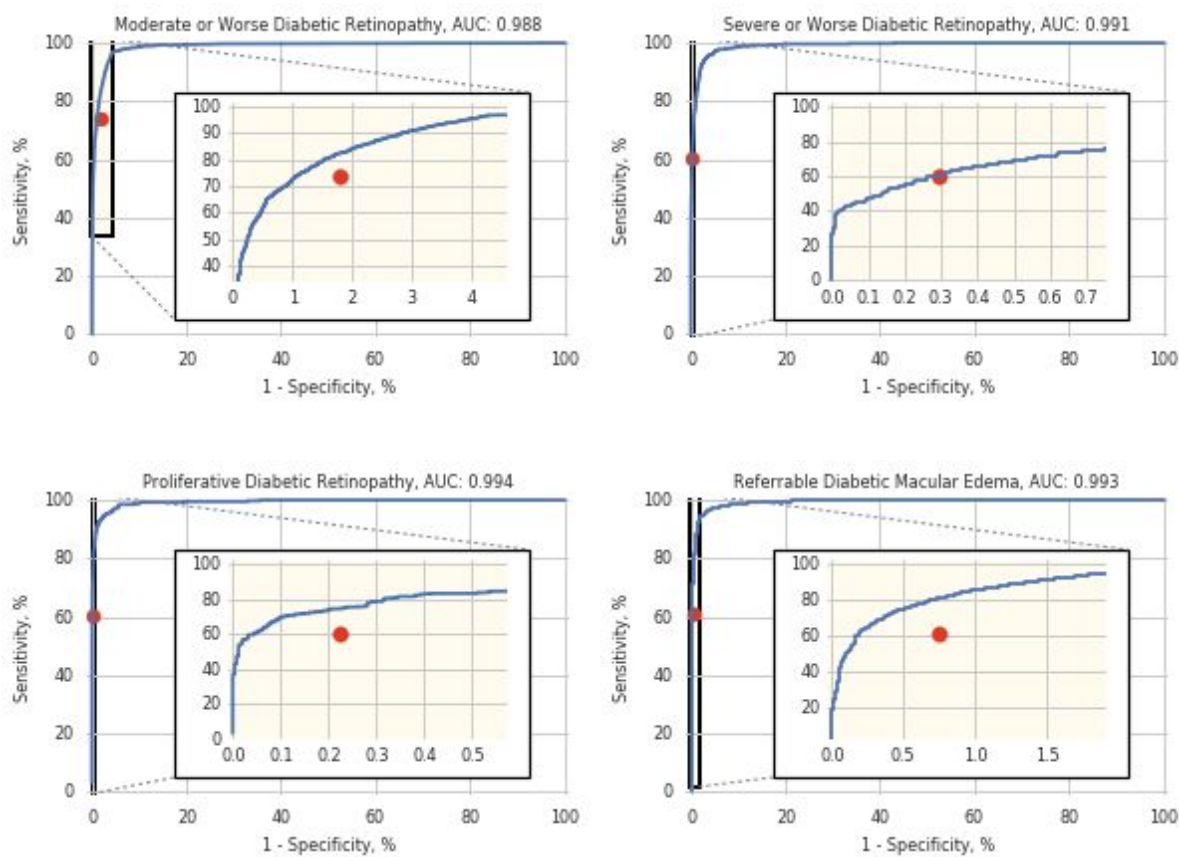

**Figure 1.** ROC curve of model performance (blue line) compared to grading by regional graders (red dot) for varying severities of DR and DME. The performance represented by the red dot is a combination of all of the grades from the regional graders on all gradable images, since regional graders only graded images from their own region.

A. Moderate or worse DR

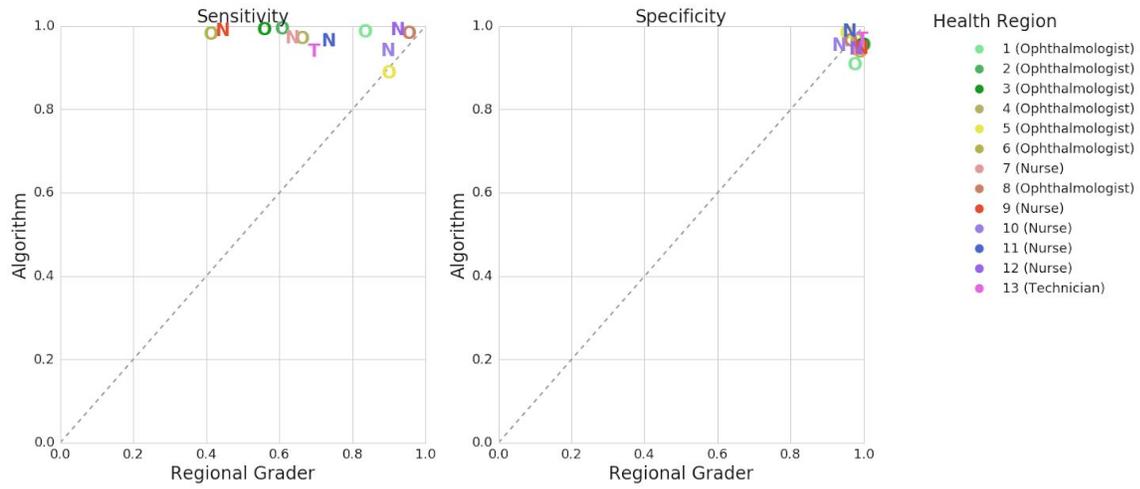

B. DME

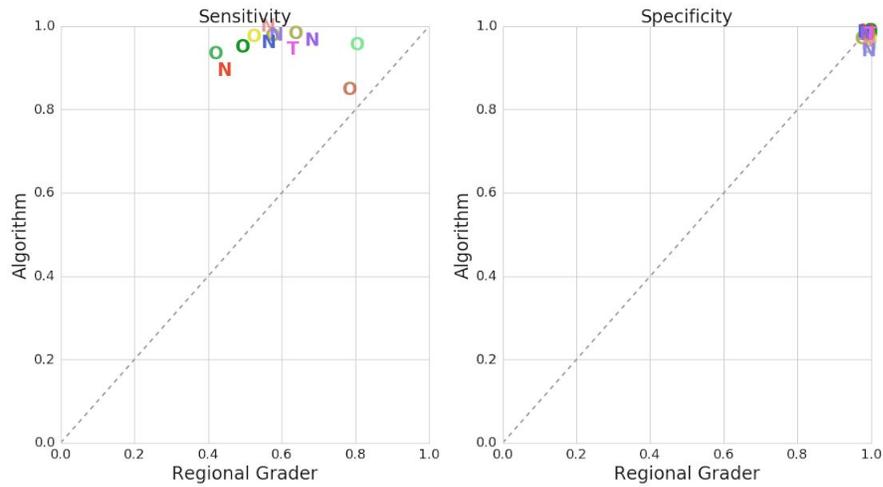

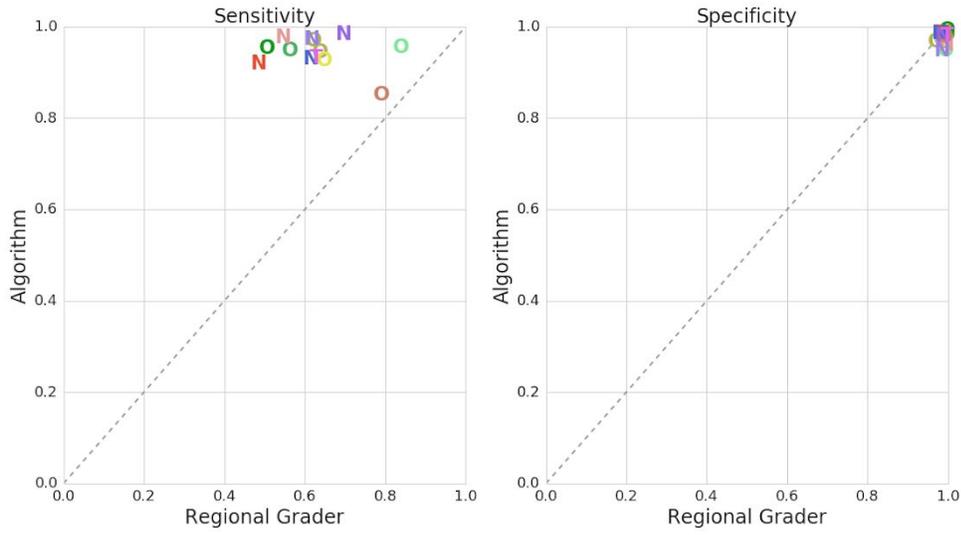

**Figure 2.** Comparison of the performance of the algorithm with regional graders for varying severities and combination of DR and DME for all gradable images. The rest of the combinations are in Supplemental Figure 1.

1.  Gulshan V, Peng L, Coram M, Stumpe MC, Wu D, Narayanaswamy A, et al. Development and Validation of a Deep Learning Algorithm for Detection of Diabetic Retinopathy in Retinal Fundus Photographs. JAMA. 2016 Dec 13;316(22):2402–10.

2.  Krause J, Gulshan V, Rahimy E, Karth P, Widner K, Corrado GS, et al. Grader Variability and the Importance of Reference Standards for Evaluating Machine Learning Models for Diabetic Retinopathy. Ophthalmology. 2018 Aug;125(8):1264–72.

3.  Ting DSW, Cheung CY-L, Lim G, Tan GSW, Quang ND, Gan A, et al. Development and Validation of a Deep Learning System for Diabetic Retinopathy and Related Eye Diseases Using Retinal Images From Multiethnic Populations With Diabetes. JAMA. 2017 Dec 12;318(22):2211–23.

4.  Jenchitr W, Hanutsaha P, Iamsirithaworn S, Parnrat U, Choosri P, Yenjitr C. The National Survey of Blindness Low Vision and Visual Impairment in Thailand 2006--2007. Thai J Pub Hlth Ophthalmol. 2007;21(1):10.

5.  Isipradit S, Sirimaharaj M, Charukamnoetkanok P, Thonginnetra O, Wongsawad W, Sathornsumetee B, et al. The First Rapid Assessment of Avoidable Blindness (RAAB) in Thailand. PLoS One [Internet]. 2014 [cited 2018 Sep 3];9(12). Available from: https://www.ncbi.nlm.nih.gov/pmc/articles/PMC4263597/

6.  Meenakshi Kumar, Sobha Sivaprasad, Shahina Pradhan, Wantanee Dangboon, Paisan Ruamvinoonsuk, Rajiv Raman. Perceived barriers for diabetic retinopathy screening by the staff involved in screening in Thailand.

7.  Ulrich Freudenstein JV. A national screening programme for diabetic retinopathy : Needs to learn the lessons of existing screening programmes. BMJ : British Medical Journal. 2001 Jul 7;323(7303):4.

8.  International Council of Ophthalmology : Resources : International Clinical Diabetic Retinopathy Disease Severity Scale, Detailed Table [Internet]. [cited 2018 Aug 24]. Available from: http://www.icoph.org/resources/45/International-Clinical-Diabetic-Retinopathy-Disease-Severity-Scale-Detailed-Table-.html

9.  Bresnick GH, Mukamel DB, Dickinson JC, Cole DR. A screening approach to the surveillance of patients with diabetes for the presence of vision-threatening retinopathy. Ophthalmology. 2000;107(1):19–24.

10. Li HK, Horton M, Bursell S-E, Cavallerano J, Zimmer-Galler I, Tennant M, et al. Telehealth Practice Recommendations for Diabetic Retinopathy, Second Edition. Telemedicine and e-Health. 2011;17(10):814–37.

11. Supapluksakul S, Ruamviboonsuk P, Chaowakul W. The prevalence of diabetic retinopathy

in Trang province determined by retinal photography and comprehensive eye examination. J Med Assoc Thai. 2008 May;91(5):716–22.

12. Szegedy C, Ioffe S, Vanhoucke V, Alemi A. Inception-v4, Inception-ResNet and the Impact of Residual Connections on Learning [Internet]. Thirty-First AAAI Conference on Artificial Intelligence. 2016 [cited 2017 Dec 13]. Available from: http://www.aaai.org/ocs/index.php/AAAI/AAAI17/paper/download/14806/14311

13. Diabetic Retinopathy Screening qualifications and training courses | City & Guilds [Internet]. [cited 2018 Sep 21]. Available from: https://www.cityandguilds.com/qualifications-and-apprenticeships/health-and-social-care/health/7360-diabetic-retinopathy-screening#tab=information

14. CPG Secretariat. Screening of Diabetic Retinopathy [Internet]. Portal Rasmi Kemeterian Kesihatan Malaysia. 2011 [cited 2018 Sep 21]. Available from: www.moh.gov.my/penerbitan/CPG2017/6601.pdf

15. Cook S E al. Quality assurance in diabetic retinal screening in South Africa. - PubMed - NCBI [Internet]. [cited 2018 Sep 21]. Available from: https://www.ncbi.nlm.nih.gov/pubmed/25363058

16. Diabetic Retinopathy PPP - Updated 2017 [Internet]. American Academy of Ophthalmology. 2017 [cited 2018 Aug 25]. Available from: https://www.aao.org/preferred-practice-pattern/diabetic-retinopathy-ppp-updated-2017

17. Chakrabarti R, Harper CA, Keeffe JE. Diabetic retinopathy management guidelines. Expert Rev Ophthalmol. 2012;7(5):417–39.

18. Wong TY, Sun J, Kawasaki R, Ruamviboonsuk P, Gupta N, Lansingh VC, et al. Guidelines on Diabetic Eye Care: The International Council of Ophthalmology Recommendations for Screening, Follow-up, Referral, and Treatment Based on Resource Settings. Ophthalmology [Internet]. 2018 May 24; Available from: http://dx.doi.org/10.1016/j.ophtha.2018.04.007

19. Scanlon PH. Screening Intervals for Diabetic Retinopathy and Implications for Care. Curr Diab Rep [Internet]. 2017 [cited 2018 Aug 25];17(10). Available from: https://www.ncbi.nlm.nih.gov/pmc/articles/PMC5585285/

20. Virgili G, Menchini F, Murro V, Peluso E, Rosa F, Casazza G. Optical coherence tomography (OCT) for detection of macular oedema in patients with diabetic retinopathy. In: Cochrane Database of Systematic Reviews. 2011.

# Supplemental Material

|  |  | Algorithm | |
|---|---|---|---|
|  |  | DR Gradable | DR Ungradable |
| Regional Grader | DR Gradable | 25,348 | 1,765 |
|  | DR Ungradable | 1,239 | 1,591 |

|  |  | Algorithm | |
|---|---|---|---|
|  |  | DME Gradable | DME Ungradable |
| Regional Grader | DME Gradable | 24,332 | 2,467 |
|  | DME Ungradable | 1,036 | 2,108 |

**Table S1.** Agreement between regional graders and the algorithm for determining image gradability for DR and DME.

|  |  | Regional Grader | |
|---|---|---|---|
|  |  | DR Gradable | DR Ungradable |
| Adjudication | DR Gradable | 56 | 155 |
|  | DR Ungradable | 547 | 224 |

|  |  | Algorithm | |
|---|---|---|---|
|  |  | DR Gradable | DR Ungradable |
| Adjudication | DR Gradable | 155 | 56 |
|  | DR Ungradable | 224 | 547 |

**Table S2**. Agreement between the adjudicated reference standard and the regional grader or the algorithm for images where the regional grader and the algorithm disagreed on DR gradability.

|  |  | Regional Grader | |
|---|---|---|---|
|  |  | DME Gradable | DME Ungradable |
| Adjudication | DME Gradable | 493 | 277 |
|  | DME Ungradable | 196 | 28 |

|  |  | Algorithm | |
|---|---|---|---|
|  |  | DME Gradable | DME Ungradable |
| Adjudication | DME Gradable | 277 | 493 |
|  | DME Ungradable | 28 | 196 |

**Table S3.** Agreement between the adjudicated reference standard and the regional grader or the algorithm for images where the regional grader and the algorithm disagreed on DME gradability.

| Region | Moderate or Worse DR | | | | | Severe or Worse DR | | | | | | Proliferative DR | | | | | | DME | | | | | |
|---|---|---|---|---|---|---|---|---|---|---|---|---|---|---|---|---|---|---|---|---|---|---|---|
| | Count | Sensitivity | | Specificity | | AUC Algorithm | Count | Sensitivity | | Specificity | | AUC Algorithm | Count | Sensitivity | | Specificity | | AUC Algorithm | Count | Sensitivity | | Specificity | | AUC Algorithm |
| | | Regional Grader | Algorithm | Regional Grader | Algorithm | | | Regional Grader | Algorithm | Regional Grader | Algorithm | | | Regional Grader | Algorithm | Regional Grader | Algorithm | | | Regional Grader | Algorithm | Regional Grader | Algorithm | |
| All | 3083 | 0.740 [0.724-0.755] | 0.970 [0.964-0.976] | 0.982 [0.980-0.984] | 0.957 [0.954-0.960] | 0.988 | 602 | 0.603 [0.563-0.642] | 0.927 [0.903-0.946] | 0.997 [0.996-0.998] | 0.976 [0.974-0.978] | 0.991 | 398 | 0.606 [0.556-0.654] | 0.719 [0.672-0.762] | 0.998 [0.997-0.998] | 0.999 [0.998-0.999] | 0.994 | 1869 | 0.613 [0.591-0.635] | 0.940 [0.928-0.950] | 0.992 [0.991-0.994] | 0.982 [0.980-0.984] | 0.993 |
| 1 | 226 | 0.836 [0.781-0.882] | 0.987 [0.962-0.997] | 0.973 [0.955-0.986] | 0.903 [0.874-0.928] | 0.982 | 56 | 0.607 [0.468-0.735] | 0.982 [0.904-1.000] | 0.997 [0.989-1.000] | 0.915 [0.891-0.935] | 0.988 | 33 | 0.788 [0.611-0.910] | 0.848 [0.681-0.949] | 0.999 [0.992-1.000] | 0.994 [0.985-0.998] | 0.993 | 133 | 0.805 [0.727-0.868] | 0.955 [0.904-0.983] | 0.993 [0.982-0.998] | 0.947 [0.925-0.964] | 0.981 |
| 2 | 140 | 0.607 [0.521-0.689] | 0.993 [0.961-1.000] | 1.000 [0.998-1.000] | 0.955 [0.944-0.965] | 0.996 | 37 | 0.649 [0.475-0.798] | 0.946 [0.818-0.993] | 0.999 [0.997-1.000] | 0.983 [0.976-0.989] | 0.997 | 26 | 0.692 [0.482-0.857] | 0.885 [0.698-0.976] | 1.000 [0.998-1.000] | 0.998 [0.995-1.000] | 0.998 | 76 | 0.421 [0.309-0.540] | 0.934 [0.853-0.978] | 0.999 [0.996-1.000] | 0.989 [0.982-0.993] | 0.995 |
| 3 | 125 | 0.560 [0.468-0.649] | 0.992 [0.956-1.000] | 0.996 [0.993-0.999] | 0.950 [0.940-0.959] | 0.991 | 13 | 0.308 [0.091-0.614] | 0.846 [0.546-0.981] | 0.999 [0.997-1.000] | 0.991 [0.986-0.995] | 0.995 | 9 | 0.444 [0.137-0.788] | 0.444 [0.137-0.788] | 0.999 [0.997-1.000] | 1.000 [0.998-1.000] | 0.991 | 79 | 0.494 [0.379-0.609] | 0.949 [0.875-0.986] | 0.998 [0.996-1.000] | 0.992 [0.988-0.996] | 0.998 |
| 4 | 101 | 0.663 [0.562-0.754] | 0.970 [0.916-0.994] | 0.980 [0.972-0.987] | 0.964 [0.952-0.973] | 0.980 | 27 | 0.630 [0.424-0.806] | 0.889 [0.708-0.976] | 0.994 [0.988-0.997] | 0.985 [0.977-0.991] | 0.978 | 16 | 0.438 [0.198-0.701] | 0.750 [0.476-0.927] | 0.995 [0.990-0.998] | 0.999 [0.995-1.000] | 0.974 | 58 | 0.638 [0.501-0.760] | 0.983 [0.908-1.000] | 0.996 [0.990-0.999] | 0.974 [0.963-0.983] | 0.993 |
| 5 | 151 | 0.901 [0.841-0.943] | 0.887 [0.826-0.933] | 0.952 [0.938-0.964] | 0.984 [0.974-0.990] | 0.988 | 59 | 0.695 [0.561-0.808] | 0.831 [0.710-0.916] | 0.997 [0.991-0.999] | 0.980 [0.970-0.987] | 0.988 | 52 | 0.731 [0.590-0.844] | 0.673 [0.529-0.797] | 0.998 [0.994-1.000] | 0.997 [0.993-0.999] | 0.992 | 82 | 0.524 [0.411-0.636] | 0.976 [0.915-0.997] | 0.998 [0.993-1.000] | 0.970 [0.959-0.980] | 0.995 |
| 6 | 254 | 0.413 [0.352-0.477] | 0.980 [0.955-0.994] | 0.991 [0.986-0.995] | 0.937 [0.925-0.947] | 0.990 | 55 | 0.782 [0.650-0.882] | 0.909 [0.800-0.970] | 0.993 [0.988-0.996] | 0.975 [0.967-0.981] | 0.990 | 38 | 0.737 [0.569-0.866] | 0.658 [0.486-0.804] | 0.996 [0.992-0.998] | 0.997 [0.994-0.999] | 0.994 | 181 | 0.575 [0.499-0.648] | 0.978 [0.944-0.994] | 0.973 [0.965-0.980] | 0.975 [0.967-0.981] | 0.995 |
| 7 | 349 | 0.636 [0.583-0.687] | 0.971 [0.948-0.986] | 0.991 [0.985-0.995] | 0.942 [0.930-0.953] | 0.987 | 94 | 0.340 [0.246-0.445] | 0.926 [0.853-0.970] | 0.999 [0.996-1.000] | 0.955 [0.944-0.963] | 0.983 | 66 | 0.258 [0.158-0.380] | 0.606 [0.478-0.724] | 1.000 [0.998-1.000] | 0.998 [0.995-0.999] | 0.986 | 236 | 0.564 [0.498-0.628] | 1.000 [0.984-1.000] | 0.998 [0.994-1.000] | 0.978 [0.970-0.985] | 0.997 |
| 8 | 659 | 0.954 [0.936-0.969] | 0.983 [0.970-0.992] | 0.961 [0.953-0.968] | 0.963 [0.956-0.969] | 0.990 | 62 | 0.645 [0.513-0.763] | 0.968 [0.888-0.996] | 0.999 [0.998-1.000] | 0.971 [0.965-0.976] | 0.993 | 41 | 0.780 [0.624-0.894] | 0.732 [0.571-0.858] | 0.999 [0.998-1.000] | 0.999 [0.997-1.000] | 0.997 | 354 | 0.785 [0.739-0.827] | 0.847 [0.806-0.883] | 0.993 [0.989-0.995] | 0.984 [0.979-0.988] | 0.984 |
| 9 | 288 | 0.444 [0.386-0.504] | 0.990 [0.970-0.998] | 0.995 [0.991-0.997] | 0.945 [0.935-0.953] | 0.991 | 62 | 0.694 [0.563-0.804] | 0.952 [0.865-0.990] | 0.993 [0.989-0.996] | 0.981 [0.975-0.986] | 0.995 | 26 | 0.577 [0.369-0.766] | 0.692 [0.482-0.857] | 0.992 [0.988-0.995] | 0.999 [0.996-1.000] | 0.994 | 198 | 0.444 [0.374-0.517] | 0.894 [0.842-0.933] | 0.989 [0.985-0.993] | 0.989 [0.985-0.993] | 0.992 |
| 10 | 342 | 0.898 [0.861-0.928] | 0.942 [0.911-0.964] | 0.943 [0.927-0.956] | 0.959 [0.945-0.970] | 0.986 | 65 | 0.569 [0.440-0.692] | 0.954 [0.871-0.990] | 0.998 [0.993-1.000] | 0.945 [0.931-0.956] | 0.987 | 49 | 0.571 [0.422-0.712] | 0.735 [0.589-0.851] | 0.998 [0.993-1.000] | 0.998 [0.995-1.000] | 0.997 | 244 | 0.586 [0.521-0.649] | 0.980 [0.953-0.993] | 0.996 [0.990-0.999] | 0.962 [0.949-0.973] | 0.993 |
| 11 | 113 | 0.735 [0.643-0.813] | 0.965 [0.912-0.990] | 0.965 [0.954-0.974] | 0.987 [0.980-0.992] | 0.996 | 21 | 0.714 [0.478-0.887] | 0.810 [0.581-0.946] | 0.994 [0.988-0.997] | 0.986 [0.979-0.992] | 0.992 | 17 | 0.824 [0.566-0.962] | 0.706 [0.440-0.897] | 0.997 [0.992-0.999] | 0.999 [0.996-1.000] | 0.994 | 76 | 0.566 [0.447-0.679] | 0.961 [0.889-0.992] | 0.983 [0.975-0.989] | 0.992 [0.986-0.996] | 0.998 |
| 12 | 119 | 0.924 [0.861-0.965] | 0.992 [0.954-1.000] | 0.977 [0.963-0.986] | 0.948 [0.930-0.963] | 0.996 | 23 | 0.522 [0.306-0.732] | 1.000 [0.852-1.000] | 0.998 [0.992-1.000] | 0.967 [0.952-0.978] | 0.996 | 11 | 0.273 [0.060-0.610] | 0.909 [0.587-0.998] | 0.999 [0.994-1.000] | 0.999 [0.994-1.000] | 0.995 | 60 | 0.683 [0.550-0.797] | 0.967 [0.885-0.996] | 0.988 [0.978-0.995] | 0.983 [0.971-0.991] | 0.996 |
| 13 | 216 | 0.694 [0.628-0.755] | 0.940 [0.899-0.968] | 0.996 [0.993-0.998] | 0.971 [0.965-0.977] | 0.978 | 28 | 0.750 [0.551-0.893] | 0.929 [0.765-0.991] | 0.999 [0.998-1.000] | 0.993 [0.989-0.995] | 0.991 | 14 | 0.786 [0.492-0.953] | 0.929 [0.661-0.998] | 0.999 [0.998-1.000] | 1.000 [0.998-1.000] | 1.000 | 92 | 0.630 [0.523-0.729] | 0.946 [0.878-0.982] | 0.997 [0.995-0.999] | 0.987 [0.982-0.990] | 0.994 |

**Table S4.** Performance of the grader compared to the algorithm by region for DR and DME at different severity thresholds for all gradable images.

| Region | DR: Quadratic Kappa | |
|---|---|---|
| | Regional Grader | Algorithm |
| All | 0.776 [0.757-0.792] | 0.846 [0.835-0.856] |
| 1 | 0.862 [0.813-0.904] | 0.856 [0.811-0.891] |
| 2 | 0.806 [0.724-0.879] | 0.852 [0.808-0.886] |
| 3 | 0.633 [0.507-0.746] | 0.745 [0.690-0.793] |
| 4 | 0.658 [0.544-0.760] | 0.820 [0.755-0.873] |
| 5 | 0.836 [0.778-0.886] | 0.853 [0.795-0.905] |
| 6 | 0.689 [0.605-0.760] | 0.785 [0.737-0.828] |
| 7 | 0.664 [0.604-0.720] | 0.850 [0.822-0.876] |
| 8 | 0.878 [0.851-0.900] | 0.862 [0.835-0.884] |
| 9 | 0.643 [0.567-0.715] | 0.823 [0.786-0.850] |
| 10 | 0.816 [0.769-0.857] | 0.872 [0.840-0.899] |
| 11 | 0.735 [0.640-0.814] | 0.864 [0.800-0.915] |
| 12 | 0.815 [0.742-0.878] | 0.866 [0.829-0.897] |
| 13 | 0.823 [0.761-0.870] | 0.830 [0.791-0.861] |

**Table S5.** Performance of the grader compared to the algorithm by region for DR. Agreement is measured in terms of quadratic weighted kappa and unweighted kappa for the 4 categories of DR (No/Mild, Moderate, Severe, Proliferative)

| Camera Make | Camera Models |
|---|---|
| 3nethra | Classic |
| Canon | CR-2 |
| Kowa | VX-10, VX-20, Nonmyd 7, Nonmyd WD, Nonmyd a-D III 8300 |
| Nidek | AFC-210, AFC-230, AFC-300 |
| Topcon | TRC NW-8 |
| Zeiss | Visucam 200 |

**Table S6.** Camera make and models used in the study. 31 total cameras (1-6 per region) were used to acquire the fundus images.

A. Severe or worse DR

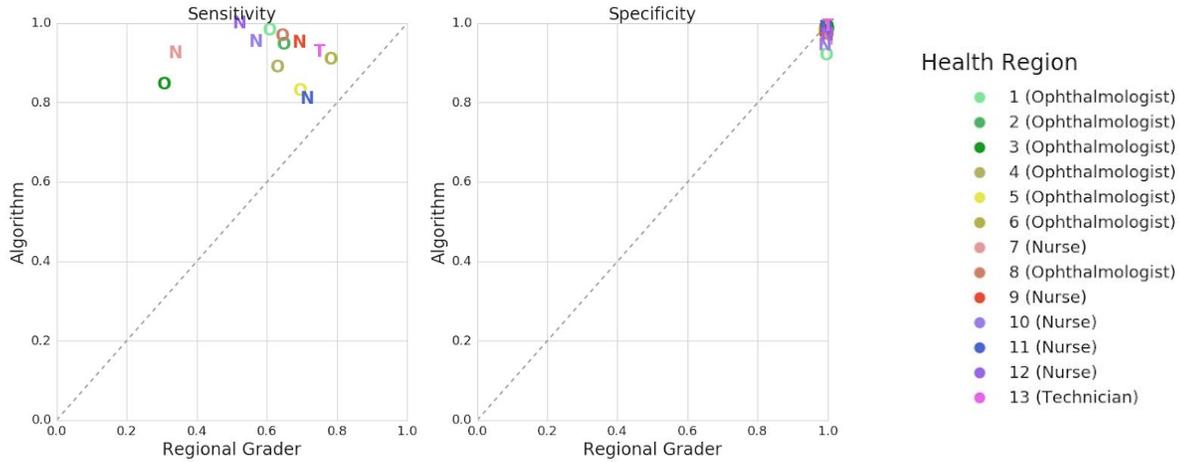

B. PDR

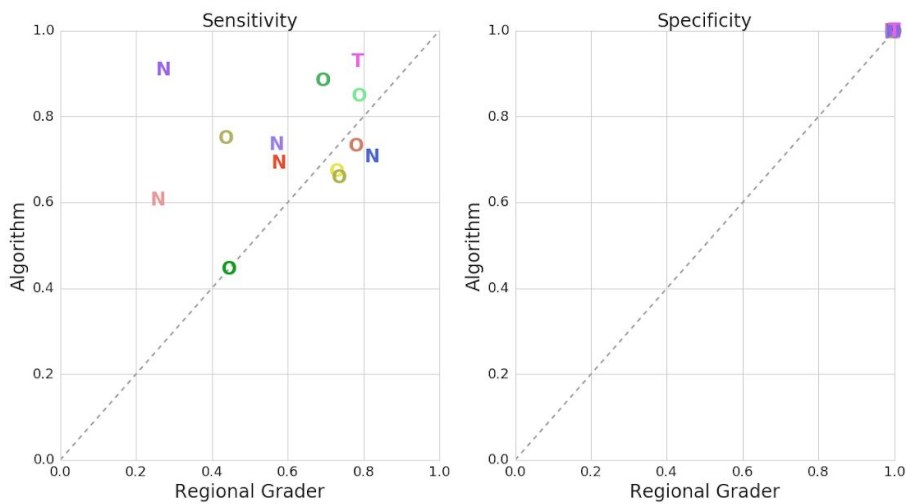

C. Moderate or worse DR and/or DME

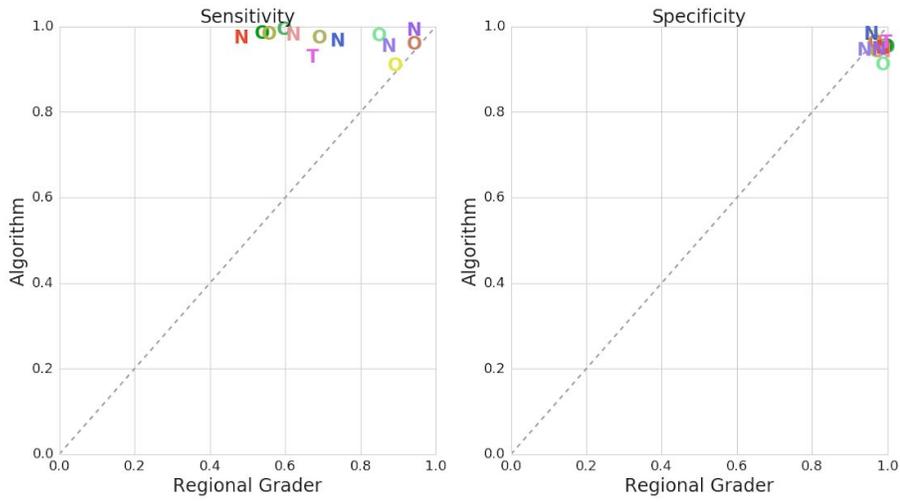

D. PDR and/or DME

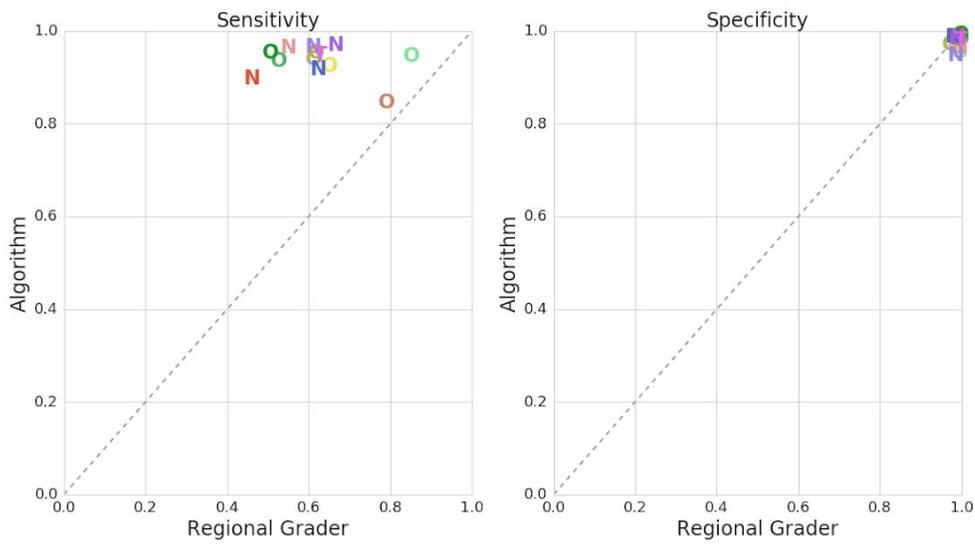

**Figure S1.** Comparison of the performance of the algorithm with regional graders for varying severities and combination of DR and DME for all gradable images. The rest of the combinations are in Figure 2 of the main manuscript.

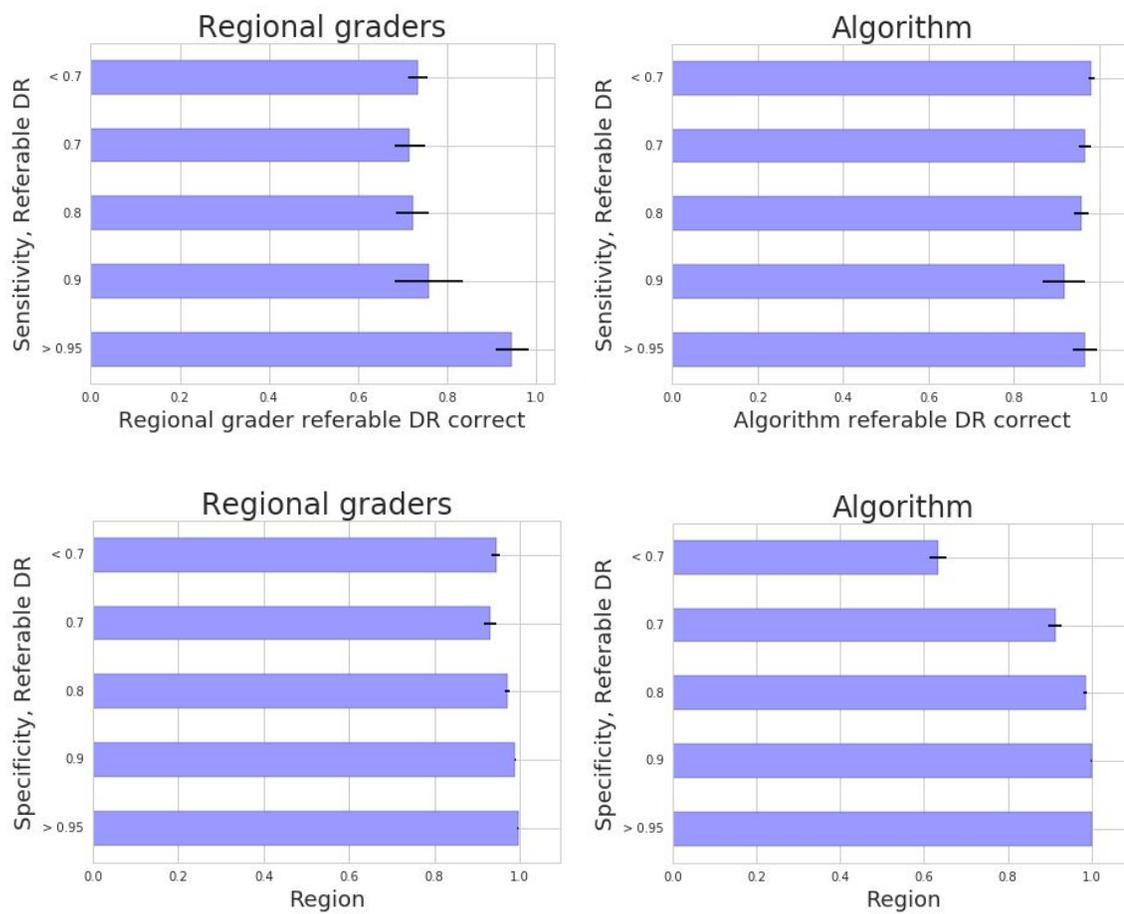

**Figure S2.** Comparison of sensitivity and specificity of regional graders and the algorithm based on the maximum confidence score of the algorithm. The algorithm is more sensitive than regional graders at all levels of confidence but less specific for low-confidence images (where the algorithm confidence scores <0.7).